\documentclass{article}

\usepackage[eandd,final]{neurips_2026}

\usepackage[utf8]{inputenc} 
\usepackage[T1]{fontenc}    
\usepackage{hyperref}       
\usepackage{url}            
\usepackage{booktabs}       
\usepackage{amsfonts}       
\usepackage{nicefrac}       
\usepackage{microtype}      
\usepackage{xcolor}         
\usepackage{tabularx}
\title{When Depth Lies: Benchmarking Vision-Language Models on Mirror-Induced RGB-D Ambiguity}

\author{%
  Hao Yin \\
  The University of Queensland, Australia \\
  \texttt{hao.yin1@uq.edu.au}
  \And
  Tianchen Guo \\
  The University of Queensland, Australia \\
  \texttt{tianchen.guo@uq.edu.au}
  \And
  Heming Du \\
  The University of Queensland, Australia \\
  \texttt{heming.du@uq.edu.au}
  \AND
  Yan Ke \\
  The University of Queensland, Australia \\
  \texttt{y.ke@uq.edu.au}
  \And
  Yanbin Liu \\
  Australian Artificial Intelligence Institute \\
  University of Technology Sydney, Australia \\
  \texttt{yanbin.liu@uts.edu.au}
  \And
  Xin Yu \\
  Australian Institute for Machine Learning \\
  Adelaide University, Australia \\
  \texttt{xin.yu@adelaide.edu.au}
}

\usepackage{microtype}  
\usepackage[table]{xcolor}
\usepackage{enumitem}
\usepackage{graphicx}
\usepackage{siunitx}
\usepackage{wrapfig}

\usepackage{color}

\begin{document}

\maketitle

\begin{abstract}
\label{sec:abs}
Vision-language models (VLMs) have made significant progress in spatial reasoning, particularly by integrating RGB images with depth data, which is often assumed to faithfully reflect physical distances. 
However, mirrors violate this assumption: they distort depth readings while reproducing object appearances, making reflections difficult to distinguish from real objects. As existing benchmarks rarely consider such reflective scenarios, VLM performance in mirror scenes remains largely unexplored.
In this work, we introduce a mirror-centric RGB-D dataset from reflective scenes to evaluate the spatial reasoning capabilities of state-of-the-art VLMs in these scenarios. Specifically, we collect 1,011 RGB-D images from 525 real-world scenes, spanning diverse indoor and outdoor environments, six reflective surface types, and both daytime and nighttime conditions. All images are annotated with reflective surface region masks and  mask annotations for both directly observed and mirror-reflected objects.
Then, we curate a benchmark named \textbf{MIRA-Bench} (Mirror-Induced Reasoning Ambiguity Benchmark), comprising 14,140 QA pairs across eight sub-tasks organized into three cognitive levels: \mbox{Reflection-aware} Perception, Spatial Reasoning, and \mbox{Scene-level} \mbox{Decision-making}. 
We evaluate state-of-the-art VLMs on MIRA-Bench and find consistent failures across tasks: models misclassify whether objects are directly observed or reflected, fail to match reflected instances to their \mbox{real-world} counterparts, and produce erroneous spatial inferences. Remarkably, providing depth data actually degrades reasoning performance, implying that VLMs cannot identify incorrect depth in mirror scenes. The code and data are available at \url{https://github.com/EvaYin-cloud/MIRA-Bench}.
\end{abstract}

\section{Introduction}
\label{sec:intro}
Vision-language models (VLMs) have made significant strides in scene understanding and spatial reasoning~\cite{spatialvlm2024, spatialrgpt2024, scenellm2024, guo2026humansspacemllmsconstruct}. 
Models such as SpatialBot~\cite{spatialbot2024} and SpatialRGPT~\cite{spatialrgpt2024} have shown that pairing RGB images with depth data substantially improves spatial reasoning. However, this success relies on an implicit premise: depth measurements accurately reflect the physical distances to observed objects. Mirrors violate this premise, as they distort depth measurements while faithfully reproducing object appearances, making reflected objects visually difficult to distinguish from real ones.
The implications for VLMs in mirror-containing scenes remain underexplored: models may treat reflected objects as physically present, infer incorrect spatial relationships, or misjudge whether a path is safe to traverse, for example by perceiving a reflected open space as walkable or overlooking collision risks from objects visible only through a mirror.

Prior work has studied mirrors and spatial reasoning separately, leaving their intersection largely unaddressed. Mirror detection and segmentation methods focus primarily on localizing mirror regions in images~\cite{mirrornet2019, pmd2020, pdnet2021, mirrorsam2_2025}, without reasoning about what is reflected or how it relates to the physical world. Meanwhile, spatial reasoning benchmarks evaluate VLMs on object relationships, distance estimation, and 3D spatial understanding~\cite{liu2023visualspatialreasoning, spatialrgpt2024, spatialbot2024, omnispatial2025}, yet generally assume that visible contents directly correspond to physical scenes. Existing RGB-D benchmarks typically treat depth values as reliable. Hence, there is no dedicated benchmark for evaluating how VLMs reason about spatial relationships in reflective scenes.

To address this gap, we built the first mirror-containing spatial reasoning benchmark, \textbf{MIRA-Bench} (\underline{M}irror-\underline{I}nduced \underline{R}easoning \underline{A}mbiguity \underline{Bench}mark), to study the spatial reasoning behaviors in reflective scenes. 
Specifically, we curate a mirror-centric RGB-D dataset consisting of 1,011 images across 525 real-world scenes from public and newly collected data.
To facilitate real-world evaluation, the dataset covers diverse environments (indoor and outdoor), reflective surface types (planar mirrors, glass surfaces, convex mirrors, vehicle side mirrors, water surfaces, and screens), and lighting conditions (daytime and nighttime). For each image, high-quality annotations are conducted, including reflective surface masks and instance masks for both directly observed and mirror-reflected objects. In addition, the depth maps allow us to assess whether depth aids mirror-containing reasoning. 

With the annotated RGB-D data, we construct 14,140 question answering (QA) pairs to establish the MIRA-Bench. Specifically, MIRA-Bench contains eight tasks categorized into three cognitive levels of gradually increasing difficulty: \textit{\mbox{Reflection-aware} Perception} (distinguishing mirror-reflected from directly observed content), \textit{Spatial Reasoning} (inferring \mbox{real-world} spatial relationships through reflections), and \textit{\mbox{Scene-level} \mbox{Decision-making}} (making scene-level judgments informed by mirror observations).
On MIRA-Bench, we evaluate ten state-of-the-art VLMs and derive two key findings. \textit{First}, current VLMs struggle across all three cognitive levels: (i) distinguishing direct and reflected observations, (ii) inferring real-world spatial relationships from reflections, (iii) making scene-level decisions. \textit{Second}, surprisingly, depth maps do not improve mirror-containing reasoning: RGB-D models perform similarly to RGB-only models, and adding depth even severely degrades collision risk assessment, as corrupted mirror depth misleads models that treat it as reliable. These results challenge the assumed reliability of depth cues in RGB-D spatial reasoning and highlight the need for reflection-aware spatial understanding.  
 
These findings demonstrate the utility of MIRA-Bench as a structured diagnostic benchmark. By combining diverse reflective scenes, paired RGB-D observations, and a three-level cognitive hierarchy, it supports fine-grained evaluation of mirror-aware spatial reasoning across perception, spatial reasoning, and scene-level decision-making, while enabling controlled comparison between RGB and RGB-D inputs. 
Our contributions are summarized as follows:
\begin{itemize}[leftmargin=1.5em]
    \item We present MIRA-Bench, the first evaluation benchmark for spatial reasoning in reflective scenes. It consists of 525 \mbox{real-world} scenes across diverse indoor and outdoor environments, and our newly collected data supplement public datasets to cover underrepresented scene types and reflective surface categories.
    \item We provide RGB-D data with fine-grained, grounded object annotations for both reflected and directly observed spaces, including reflective surface region masks, instance mask annotations and the correspondences between real and reflected objects.
    \item We conduct the first dedicated evaluation of state-of-the-art VLMs on mirror-centric spatial reasoning, revealing consistent failures across all cognitive levels, from perception through reasoning to decision-making.
    \item We demonstrate that depth information, widely shown to improve spatial reasoning in standard scenes, provides no consistent benefit in mirror scenes and can even degrade reasoning performance, challenging a foundational assumption in RGB-D spatial reasoning.
\end{itemize}
\section{Related Work}
\subsection{VLMs for Spatial Understanding}
Recent years have witnessed rapid advances in the spatial understanding capabilities of vision-language models~\cite{liu2025spatial,zheng2025multimodal,spatialvlm2024,shiri2024empirical,spatialrgpt2024,spatialbot2024,wang2025n3d,liu2025ssr,zhu2026video,zhang20263dide3dimplicitdepth,zeng2024perceivereflectplandesigning}. Early models such as LLaVA~\cite{llava2023} and BLIP-2~\cite{blip2_2023} established foundational paradigms for vision-language alignment. SpatialVLM~\cite{spatialvlm2024} endows VLMs with quantitative spatial reasoning by training on internet-scale 3D spatial QA data in metric space. SpatialRGPT~\cite{spatialrgpt2024} extends spatial reasoning to region-level understanding by integrating depth maps for grounded 3D relationship inference. SpatialBot~\cite{spatialbot2024} directly incorporates RGB-D image pairs as input, enabling depth-aware spatial understanding without relying on external depth estimators.

N3D-VLM~\cite{wang2025n3d} integrates native 3D object perception with 3D-aware visual reasoning to achieve both 3D grounding and interpretable spatial understanding. SSR~\cite{liu2025ssr} transforms raw depth into structured, interpretable textual rationales. SD-VLM~\cite{chen2025sd} introduces depth positional encoding to improve spatial awareness in VLMs. A common thread across these advances is the use of depth as a reliable source of geometric information.

Alongside model advances, a growing number of benchmarks have been proposed to evaluate these capabilities. VSR~\cite{liu2023visualspatialreasoning} targets 2D spatial relationship classification between object pairs, while a broader set of benchmarks~\cite{cvbench2024,blink2024,guo2025plnet,guo2026beyond,guo2026ssmnbench,yeh2026seeing,fu2025mmecomprehensiveevaluationbenchmark,liu2024mmbench,du2026sameperson,shen2026fingercap,ke2026ivqald,qi2026topologyunified2dposeestimation} assess broader perceptual and spatial skills across diverse visual tasks. OmniSpatial~\cite{omnispatial2025}, grounded in cognitive psychology, extends evaluation beyond basic spatial relations to higher-order tasks such as dynamic reasoning and perspective-taking. These efforts have substantially advanced the evaluation of spatial reasoning in VLMs. However, they all evaluate spatial reasoning in standard scenes where objects are directly observed and depth values are reliable. None addresses scenarios where reflective surfaces render depth unreliable for spatial reasoning.

\subsection{Mirror Understanding in Computer Vision}
Mirrors pose a well-known challenge in computer vision, as existing systems may be confused by reflected content inside mirrors, leading to severe performance degradation~\cite{mirrornet2019}. This challenge is twofold: mirrors reproduce the visual appearance of surrounding objects, making it hard to separate real from reflected content, and they distort depth measurements, since sensors report the round-trip light path through the mirror rather than the true distance to reflected objects. MirrorNet~\cite{mirrornet2019} first addresses mirror segmentation as a computational problem by modeling contextual contrast between mirror interiors and surroundings. PMD~\cite{pmd2020} introduces a progressive approach to refine segmentation accuracy. PDNet~\cite{pdnet2021} incorporates depth cues to assist mirror boundary delineation. On the dataset side, RGBD-Mirror~\cite{pdnet2021} provides an RGB-D mirror segmentation dataset with paired depth maps. DVMD~\cite{dvmd2023} extends mirror detection to the video domain. More recently, MirrorSAM2~\cite{mirrorsam2_2025} adapts the SAM2 architecture~\cite{sam2} for video mirror segmentation with depth perception.
These efforts have effectively advanced the ability to localize mirror regions. However, they primarily address a perceptual task: determining \textit{where} a mirror is. It remains an open question how VLMs reason about spatial relationships in mirror scenes, where both RGB appearance and depth cues can be unreliable.

\subsection{Visual Referring in VLM Evaluation}
In region-specific VLM evaluation, how target objects are referred to has been shown to significantly affect model performance~\cite{yu2016modeling,xiao2025towards,ding2025multimodal,peng2023kosmos}. VRPTEST~\cite{vrptest2023} tests 12 visual referring prompt strategies and found that the choice of strategy alone can cause accuracy variations of up to 17.5\%. VP-Bench~\cite{vpbench2025} evaluates eight visual prompt types, including bounding boxes, masks, and arrows, revealing that models recognize regular prompts such as bounding boxes more reliably than irregular ones such as scribbles. Furthermore, recent work~\cite{fragile2025} demonstrates that even minor changes in marker style, such as color or thickness, can cause accuracy fluctuations exceeding 20\% and even reverse model rankings. These findings highlight that object-referring design is a non-trivial confounding factor in VLM evaluation. Motivated by this, MIRA-Bench adopts five object-referencing protocols to ensure that our findings are robust across interaction paradigms.      
\section{MIRA-Bench}
\label{sec:_bench}
\begin{figure*}[t]
\centering
\includegraphics[width=\textwidth]{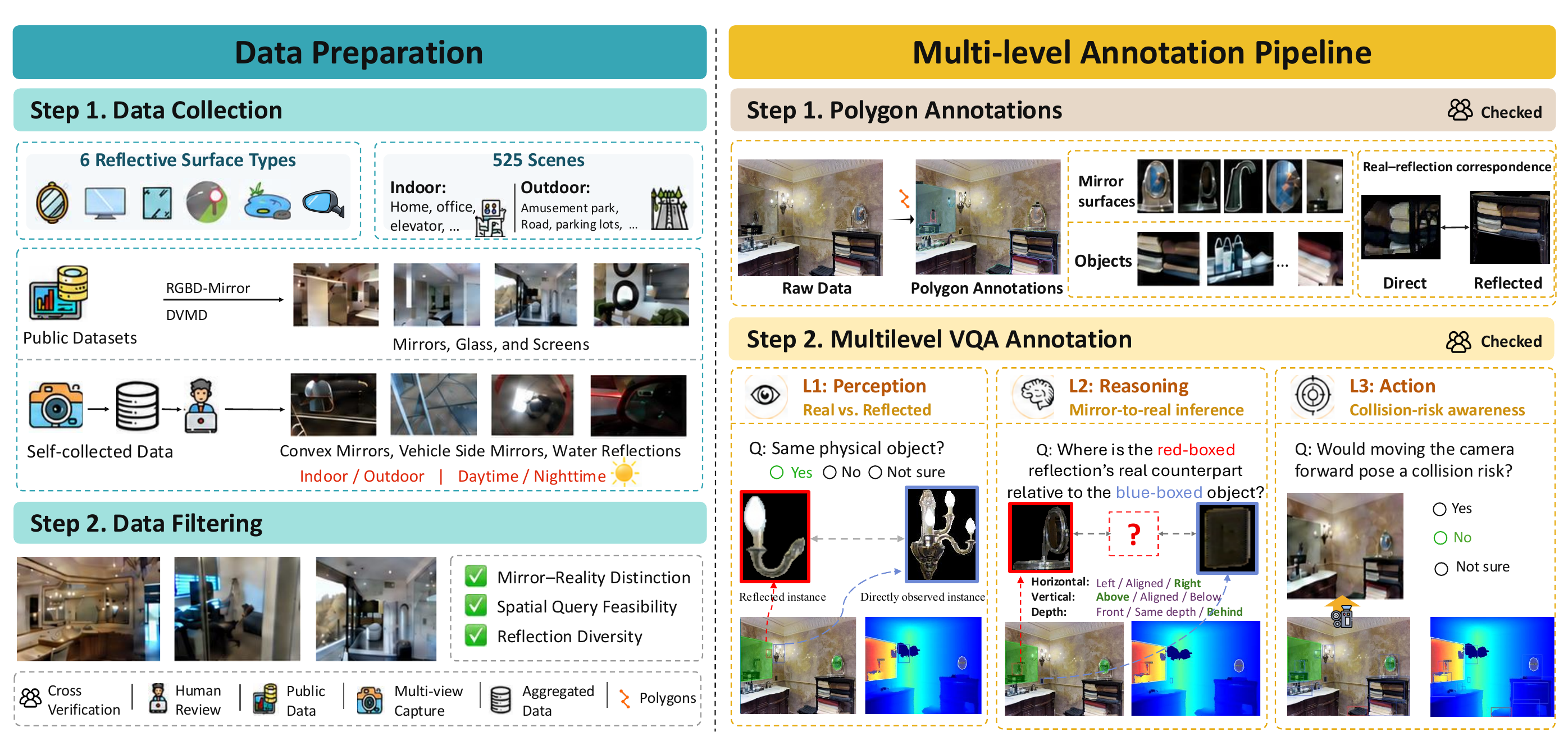}
\caption{\textbf{Overview of the MIRA-Bench construction pipeline}. Left: data collection from public datasets and newly captured scenes, followed by filtering and selection. Right: annotation pipeline combining reflective surface mask labeling, object instance polygon annotation, and QA pair generation with multi-stage quality control.}
\label{fig:pipeline}
\end{figure*}

To rigorously evaluate VLM spatial reasoning in the presence of reflective surfaces, we introduce \textbf{MIRA-Bench}. 
It is designed to assess VLMs' geometric correspondence understanding and spatial comprehension capabilities especially when depth information is distorted and RGB appearance is reflected by mirrors. The overall data collection and annotation pipeline is illustrated in Figure~\ref{fig:pipeline}. Representative QA examples are provided in
Appendix~\ref{app:qa_construction}.

\subsection{Data Source and Collection}
\label{subsec:data_source}
MIRA-Bench comprises a highly curated set of 1,011 RGB-D images distributed across 525 distinct scenes. To ensure robust domain coverage and structural diversity, the dataset is aggregated from three complementary sources, blending established public datasets with our own targeted data collection.

\textbf{Data Sources.} 
The foundation of our benchmark relies on a synergistic combination of existing public data and newly collected RGB-D images. Specifically, we select 517 high-quality images across 406 scenes from RGBD-Mirror~\cite{pdnet2021}, a public RGB-D mirror segmentation dataset. To incorporate temporally correlated reflective phenomena, we extract 64 representative keyframes spanning 34 scenes from the DVMD~\cite{dvmd2023} video-based mirror detection dataset. Furthermore, to bridge existing domain gaps and introduce more complex environments, we newly capture 430 images across 85 scenes using an Intel RealSense D435 depth camera, a widely adopted and reliable RGB-D sensor.

\textbf{Task-Driven Scenes.} 
Traditional datasets often define a scene strictly by physical geography, meaning all images captured within a single room constitute one scene. 
In contrast, MIRA-Bench defines a scene by its reflective context. If two images are captured within the same physical room but feature different reflective surfaces, or if extreme viewpoint shifts cause the same surface to reveal fundamentally different reflected content, they are classified as distinct scenes. Thus, we define these scenes as task-driven, as distinct reflective contexts inherently generate unique and independent spatial reasoning challenges.

\textbf{Addressing Domain Gaps via Data Collection.} 
Existing RGB-D mirror-centric datasets are heavily biased toward indoor, planar mirrors. However, real-world reflective phenomena are more complex. Outdoor environments frequently feature convex mirrors in parking facilities or on roadsides, vehicle side mirrors, glass facades and water surfaces. To mitigate this distributional bias, our data collection explicitly targets these underrepresented, non-planar, and outdoor scenarios. Furthermore, we deliberately capture scenes across diverse lighting conditions, including both daytime and nighttime environments, to rigorously test model reasoning robustness. For each physical setup, we capture multiple images from varying distances and viewing angles. Because the virtual frustum visible through a reflective surface is highly sensitive to the observer's position, this multi-view approach yields a rich diversity of spatial reasoning paradigms from a single physical locale.

\textbf{Rigorous Selection Criteria.} 
To guarantee both the high quality of annotations and the analytical validity of the benchmark, every image must satisfy three strict inclusion criteria. First, the reflective surface must be clearly distinguishable within the image context. Images where the entire field of view is occupied by reflected content are strictly excluded as it is impossible even for human annotators to differentiate the virtual world from the real world. Second, the scene must contain a sufficient density of objects situated in both the directly observable space and the reflected space, which is a prerequisite for constructing complex, meaningful spatial relationship queries. Third, for subsets of images captured from multiple viewpoints within the same physical scene, the reflective surface should present distinctly different reflected content. This ensures that each image contributes a unique set of spatial relationships rather than redundant data.

\subsection{Annotation Pipeline}
\label{subsec:annotation_pipeline}

To facilitate a targeted evaluation of reasoning capabilities, each image in MIRA-Bench is annotated comprehensively for two major components: reflective surface regions, and specific object instances and their reflected images. Details of the annotation team, training, and annotation interfaces
are provided in Appendix~\ref{app:annotators_training}.
\textbf{Reflective Surface Delineation.} 
We construct high-fidelity, region-wise masks delineating the boundaries of all reflective surfaces within each image. By explicitly providing these reflective boundaries to the models during evaluation, we eliminate mirror detection errors as a confounding variable. This allows models to focus on understanding the semantics and geometry of reflected and physical content, rather than struggling to localize mirrors.

\textbf{Class-Agnostic Instance Masking.} 
We provide polygon-based instance masks for individual objects, encompassing both directly observed items and their mirror-reflected counterparts, and assign a unique numeric identifier to each. These identifiers do not contain any semantic class information and spatial priors. They do not indicate whether an instance exists in the physical space or the virtual reflected space. 
The metadata, including object masks, object presence in mirrors or the physical world, and object correspondences, are also annotated and then used for question–answer pair generation.
The polygon annotations establish unambiguous object references in our query formulations, supporting the various object-referencing protocols detailed in Section~\ref{sec:referring}.

\textbf{Depth Map Assisted Geometric Grounding.} 
Every RGB image in the benchmark is paired with a corresponding metric depth map. In standard 2D image analysis, spatial relationships along the depth axis are typically inferred using visual heuristics, such as relative object scale and vertical positioning within the frame. However, in complex 3D environments distorted by reflections, these heuristics may fail; a nearby object may simply be physically smaller than a distant one. 
Depth maps help to resolve this inherent ambiguity in 2D projection by providing metric depth values, serving two distinct roles within MIRA-Bench. First, they provide the authoritative geometric grounding necessary for human annotators to derive ground-truth answers, particularly for ambiguous front-to-back spatial relationships. Second, during inference, depth maps can be optionally provided to the VLMs as a supplementary modality. This allows us to systematically investigate whether explicit geometric priors fundamentally enhance a model's capacity for mirror-aware spatial reasoning. 

\textbf{Quality control.} We apply a multi-stage quality assurance protocol across all annotation types.
For reflective surface masks and object instance polygons, each batch of 100 images is completed by one annotator. We randomly sample 20\% of annotations from each batch for independent re-annotation by a second annotator. Note that the annotations are not provided to the second annotator. We compute the mean Intersection-over-Union (mIoU) between the two sets; if mIoU falls below 85\%, the entire batch is reassigned to a third annotator for re-annotation. Further details of image annotation and quality control are provided
in Appendix~\ref{app:image_annotation}.
For QA pairs, all the generated questions undergo cross-verification by two independent reviewers, who check question phrasing, answer correctness, and option plausibility. Disagreements are resolved by a third reviewer. Spatial relationship answers can be additionally verified using depth maps.
      
\subsection{Task Taxonomy}
\label{sec:task}
Based on the annotations described above, we design a comprehensive QA taxonomy. It is organized primarily by cognitive complexity into three levels, which comprise eight sub-tasks.

\textbf{\mbox{Reflection-aware} Perception} evaluates whether a model can correctly distinguish between directly observed and mirror-reflected content. It comprises two sub-tasks. \textit{Visibility classification} asks the model to determine, for a given object instance, whether it is observed only through a reflection, only in the directly observed space, or in both. \textit{Mirror correspondence} tests a model's ability to establish identity mappings across reflected and physical spaces, including identifying which \mbox{real-world} instance corresponds to a given reflected instance, and vice versa.

\textbf{Spatial Reasoning} evaluates whether a model can infer \mbox{real-world} spatial relationships between reflected and directly observed spaces. It comprises three sub-tasks. \textit{Mirror-to-real spatial inference} requires the model to determine the \mbox{real-world} spatial relationship between instances observed through reflections, along horizontal, vertical, and depth axes. \textit{Counting-based spatial reasoning} asks how many objects lie in a given direction relative to a specified instance, requiring integration of spatial information from both reflected and directly observed regions. \textit{Mirror-provided attribute completion} tests whether the model can recognize that a reflection reveals additional information about an object, such as its back surface, that is not visible from the direct viewpoint.

\textbf{\mbox{Scene-level} \mbox{Decision-making}} evaluates whether a model can translate mirror-aware spatial understanding into practical scene-level judgments. It comprises three sub-tasks. \textit{Reachability} asks whether a specified object is reachable from the current camera position. \textit{Next-step action selection} asks a model to choose the best movement action to approach a target object or to better observe an object which is currently visible only through  reflection. \textit{Collision risk awareness} requires a model to assess whether moving in a given direction would pose a collision risk, based on the full scene understanding including information provided by reflective surfaces.

Ground-truth answers for reflection-aware perception are generated
automatically from the annotated visibility labels and real--reflection
correspondences. For spatial reasoning and decision making at the scene
level, answers are manually annotated through visual inspection, with
depth maps used as auxiliary evidence where applicable. Distractor choices are generated either from existing annotations or by constructing variants that follow a similar form to the ground-truth answer, ensuring that all options are visually grounded and contextually plausible. To ensure annotation quality, annotators manually verify each QA pair and its answer choices, confirming that exactly one option is correct while the remaining options serve as plausible distractors. Appendix~\ref{app:qa_construction} details the construction and
verification of QA pairs, including question templates and answer options.

\subsection{Object-Referencing Protocols}
\label{sec:referring}

Prior work has shown that how target objects are referred to can significantly affect VLM performance, with accuracy variations exceeding 17\% depending on the referring strategy~\cite{vrptest2023, vpbench2025, fragile2025}. To ensure that our findings are robust and not biased to a particular referencing method, we design five object-referencing protocols along two modalities. For textual referring, we provide either (1) bounding box coordinates alone, or (2) bounding box coordinates together with a natural language description of the object. For visual referring, we provide either (3) a bounding box drawn directly on the image, or (4) a binary polygon mask of the target object. Furthermore, we combine (5) visual and textual cues by pairing a visual bounding box with a natural language description. Illustrative examples of the five protocols are provided in
Appendix~\ref{app:qa_construction} (Figure~\ref{fig:referring_protocols}).
Note that not all sub-tasks require object-specific referencing. For instance, collision risk awareness queries a movement direction rather than a specific object, and thus does not require object references. Therefore, the number of applicable referencing protocols varies across sub-tasks, yielding 14,140 QA pairs in total.

\subsection{Dataset Statistics}
\label{sec:stats}

\begin{figure}[t]
\centering
\includegraphics[width=\textwidth]{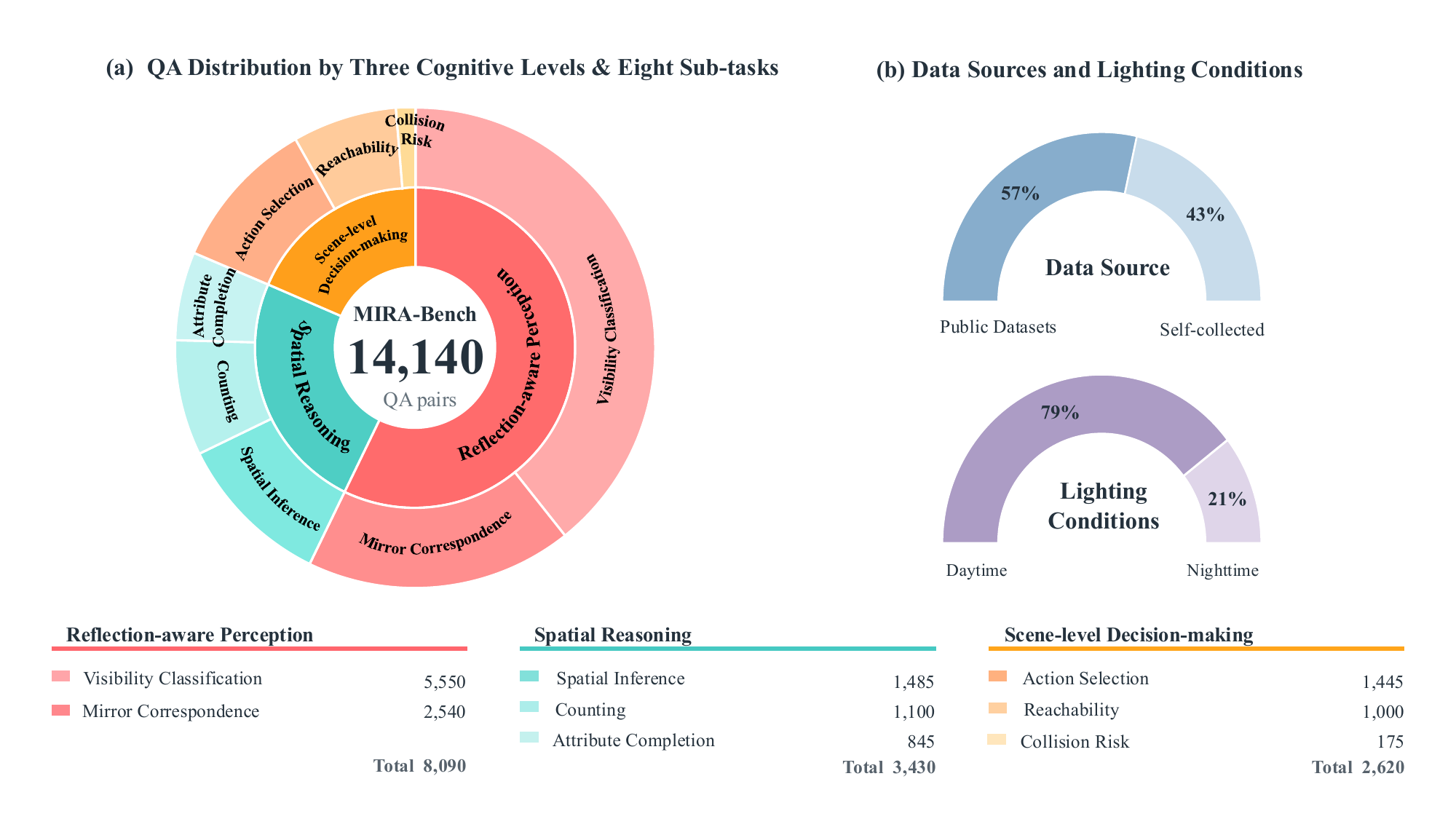}
\caption{Statistics of MIRA-Bench.
(a) Distribution of QA pairs across three cognitive levels
and eight subtasks.
(b) Distribution of images by data source and scenes by
lighting condition (daytime versus nighttime).}
\label{fig:statistics}
\end{figure}

We summarize the key statistics of MIRA-Bench in Figure~\ref{fig:statistics}. 

\textbf{Data composition.} MIRA-Bench consists of 1,011 RGBD images spanning 525 scenes, compiled from three sources: RGBD-Mirror~\cite{pdnet2021} (406 scenes, 517 images), DVMD~\cite{dvmd2023} (34 scenes, 64 images), and our newly collected dataset (85 scenes, 430 images). While public datasets provide the majority of unique scenes, our novel data collection accounts for over 42\% of the total images due to an extensive multi-viewpoint capture strategy. Figure~\ref{fig:statistics} (b) details the dataset's overall statistics, including data source and lighting conditions. Across the six reflective surface categories, planar mirrors and glass are the most prevalent, whereas convex mirrors, vehicle side mirrors, and water surfaces are predominantly introduced by our newly captured data.
 
\textbf{QA composition.} MIRA-Bench contains 14,140 QA pairs across all the images. The distribution of QA types across three cognitive levels and eight sub-tasks is shown in Figure~\ref{fig:statistics} (a). Reflection-aware Perception accounts for 8,090 QA pairs (57\%), Spatial Reasoning for 3,430 (24\%), and \mbox{Scene-level} \mbox{Decision-making} for 2,620 (19\%). 
 
\section{Experiments}
\subsection{Evaluated Models}
\label{sec:models}

We evaluate ten state-of-the-art VLMs on MIRA-Bench, covering both proprietary and open-source models. Proprietary models include GPT-5.4~\cite{gpt54_2026}, Gemini-2.5-Pro~\cite{gemini2025}, Gemini-2.5-Flash~\cite{gemini2025} and Gemini-3.8-Flash~\cite{google2026gemini38flash}. The open-source set includes three Qwen-family models, namely Qwen3.5-9B~\cite{qwen3.5}, Qwen2.5-VL-7B~\cite{bai2025qwen25vltechnicalreport}, and Qwen3-VL-8B~\cite{bai2025qwen3}, as well as LLaVA-OneVision-7B~\cite{li2024llavaonevisioneasyvisualtask}, InternVL3-8B~\cite{zhu2025internvl3exploringadvancedtraining}, and DeepSeek-VL2-Tiny~\cite{deepseekvl2_2024}. This selection spans a range of model scales and architectures that are broadly representative of current VLMs.

We additionally report two reference baselines. \textbf{Random Guess} is computed analytically from the number of answer choices per question, providing a chance-level lower bound that accounts for differences in question format across sub-tasks. \textbf{Human Performance} is measured by having human annotators answer a representative subset of the benchmark under the same visual conditions as the models, establishing an empirical upper bound for each sub-task.

\subsection{Evaluation Metrics}
\label{sec:metrics}
We adopt exact accuracy as the primary metric, defined as the percentage of questions where the predicted answer match the ground truth exactly. Evaluation details vary by question type. For multiple-choice questions, exact accuracy requires the predicted answer set to exactly match the gold answer set completely. For spatial 3D axes questions, partial accuracy is computed as the fraction of correctly predicted axes (horizontal, vertical, depth) among those present in the ground truth. For visibility classification, we employ a joint scoring strategy: each annotated object generates a triplet of related questions (vc\_001, vc\_002, vc\_003), and a triplet is scored as correct only if all three are answered correctly. We also report a random-guessing baseline for each sub-task for comparison. Performance within each cognitive level is summarized using the weighted average accuracy (W-Avg), defined as:
\[
\mathrm{W\text{-}Avg}_{l}
=
\frac{\sum_{t\in\mathcal{T}_{l}} n_t\,\mathrm{Acc}_t}
{\sum_{t\in\mathcal{T}_{l}} n_t},
\]
where \(\mathcal{T}_l\) denotes the sub-tasks in cognitive level \(l\), \(n_t\) is the number of questions in sub-task \(t\), and \(\mathrm{Acc}_t\) is its accuracy. Overall accuracy uses the same weighting across all eight sub-tasks.

\subsection{Implementation Details}
\label{sec:impl}

For the main evaluation, all models are evaluated using RGB images
and their corresponding depth maps, with mirror region masks
overlaid on both. We set the temperature to 0.0 and maximum output tokens to 2,048 for all models. Each model is prompted to return a single valid JSON object containing three fields: \texttt{final\_answer}, \texttt{brief\_explanation}, and \texttt{confidence}. If the initial response fails to parse or produces an invalid answer format, a repair pass is attempted by re-prompting the model with the malformed output. Each question is retried up to two times upon API failure. Proprietary models are accessed via their respective APIs, while open-source models are deployed locally.

\subsection{Main Results}
Table~\ref{tab:main_results} presents results across all models and sub-tasks. Human annotators achieve 85.0--90.7\% overall accuracy, while the best proprietary model, Gemini-3.8-Flash, reaches only 40.1\%, and the best open-source model, Qwen3.5-9B, reaches 30.4\%. Proprietary models generally outperform open-source ones, but both fall far short of human performance across all three cognitive levels.
 
\textbf{Reflection-aware Perception.} Models achieve moderate performance on mirror correspondence (MC), which involves matching individual instances across reflected and directly observed spaces. However, most models obtain
lower scores on visibility classification (VC), which asks whether each object is seen only through a reflection, only directly, or in both, and requires all three related questions about the same object to be answered correctly. This gap indicates that models can handle individual matching to some extent, but lack the consistency needed to make stable, repeated judgments about the same object, revealing an inherent instability in their perception of reflected scenes.
 
\textbf{Spatial Reasoning.}
Models struggle with mirror-to-real spatial inference (MR), and
difficulties also persist in counting-based spatial reasoning (CT).
Both tasks require models to infer physical spatial relationships
from reflected observations and integrate this information with
directly observed content. The limited performance suggests that
current VLMs have difficulty accounting for the geometry of reflection
when reasoning about object locations and relationships in the
underlying physical scene.
 
\textbf{Scene-level Decision-making.} Most models struggle with action selection (AS), while collision
risk (CR) assessment remains substantially below human performance
despite variation across models. These tasks require models to use
information from both reflected and directly observed regions to
assess feasible movements and potential obstacles. The observed
difficulties suggest that current VLMs cannot yet consistently
translate mirror-aware spatial understanding into reliable decisions
about movement and safety. Appendix~\ref{app:cross_level} examines how errors in reachability
and action selection are associated with errors in perception
and spatial reasoning.

\begin{table}[t]
\centering
\caption{Evaluation results on MIRA-Bench. We report exact accuracy (\%) for each sub-task. VC: Visibility Classification (joint); MC: Mirror Correspondence; MR: Mirror To Real Spatial Inference; CT: Counting; MPAC: Mirror Provided Attribute Completion; RE: Reachability; AS: Next Step Action Selection; CR: Collision Risk. W-Avg: weighted average within each cognitive level, weighted by the number of questions per sub-task. Overall: weighted average across all sub-tasks.}
\label{tab:main_results}
\fontsize{8.5}{10}\selectfont
\setlength{\tabcolsep}{2.5pt}
\renewcommand{\arraystretch}{1.0}

\begin{tabular*}{\linewidth}{@{\extracolsep{\fill}}l ccc cccc cccc c@{}}
\toprule
 & \multicolumn{3}{c}{Refl.-aware Perception} & \multicolumn{4}{c}{Spatial Reasoning} & \multicolumn{4}{c}{Scene-level Decision-making} & \\
\cmidrule(lr){2-4} \cmidrule(lr){5-8} \cmidrule(lr){9-12}

Model & VC & MC & W-Avg & MR & CT & MPAC & W-Avg & RE & AS & CR & W-Avg & Overall \\
\midrule
\rowcolor{gray!15} \multicolumn{13}{l}{\textit{Baselines}} \\
\cmidrule{1-13}
Random Guess            & 12.3 & 15.0 & 13.1 & 3.7 & 2.5 & 25.0 & 8.6 & 33.3 & 16.7 & 33.3 & 24.1 & 14.1 \\
Human (RGB)  & 85.1 & 82.3 & 84.2 & 79.3 & 78.0 & 90.8 & 81.7 & 92.2 & 91.5 & 90.2 & 91.7 & 85.0 \\
Human (RGB-D)& 92.3 & 90.5 & 91.7 & 85.0 & 80.2 & 92.1 & 85.2 & 94.4 & 94.9 & 91.5 & 94.5 & 90.7 \\
\midrule
\rowcolor{gray!15} \multicolumn{13}{l}{\textit{Proprietary Models}} \\
\cmidrule{1-13}
GPT-5.4                  & 30.5 & 42.7 & 34.3 & 17.7 & 22.7 & 48.2 & 26.8 & 64.3 & 9.7 & 54.3 & 33.5 & 32.3 \\
Gemini-2.5-Pro           & 31.7 & 39.3 & 34.1 & \textbf{22.3} & 17.2 & 51.4 & 27.8 & 65.3 & 10.9 & 58.9 & 34.8 & 32.7 \\
Gemini-2.5-Flash         & 27.8 & 41.0 & 31.9 & 19.1 & 17.5 & 50.7 & 26.4 & 62.1 & 12.3 & \textbf{61.1} & 34.6 & 31.1 \\
Gemini-3.8-Flash         & \textbf{37.4} & \textbf{48.4} & \textbf{40.8} & 16.8 & 18.4 & \textbf{67.7} & \textbf{29.9} & \textbf{66.9} & \textbf{39.6} & 60.0 & \textbf{51.4} & \textbf{40.1} \\
\midrule
\rowcolor{gray!15} \multicolumn{13}{l}{\textit{Open-source Models}} \\
\cmidrule{1-13}
Qwen3.5-9B              & 32.5 & 42.4 & 35.6 & 2.9 & 17.2 & 49.8 & 19.0 & 56.0 & 9.8 & 38.3 & 29.3 & 30.4 \\
Qwen3-VL-8B             & 28.1 & 25.6 & 27.3 & 3.2 & 20.3 & 32.5 & 15.9 & 59.7 & 12.0 & 25.1 & 31.1 & 25.2 \\
Qwen2.5-VL-7B           & 21.7 & 24.6 & 22.6 & 0.4 & 21.5 & 38.1 & 16.5 & 57.0 & 15.0 & 21.7 & 31.5 & 22.8 \\
LLaVA-OV-7B             & 15.1 & 17.7 & 15.9 & 4.4 & 23.1 & 34.6 & 17.8 & 54.7 & 13.1 & 26.9 & 29.9 & 19.0 \\
InternVL3-8B            & 21.6 & 19.1 & 20.8 & 2.4 & 28.2 & 42.5 & 20.5 & 58.6 & 14.3 & 30.9 & 32.3 & 22.9 \\
DeepSeek-VL2-Tiny       & 13.2 & 17.9 & 14.7 & 0.9 & \textbf{31.0} & 27.2 & 17.1 & 57.2 & 3.4 & 23.4 & 25.3 & 17.2 \\
\bottomrule
\end{tabular*}
\end{table}

\subsection{Modality Analysis}
\label{sec:input}
Depth information has been shown to improve spatial reasoning in standard scenes, and mirror masks provide explicit reflective boundaries that could guide model attention. To investigate whether these additional signals help in mirror scenes, we compare four input configurations on Qwen3.5-9B: raw RGB, RGB with depth (RGB-D), RGB with a mirror-region mask (RGB-M), and RGB with both (RGB-DM). As shown in Table~\ref{tab:input_modality}, the overall accuracy varies within a narrow range of 29.6--31.3\% across all four modes, indicating that neither depth nor the mirror mask substantially improves mirror-aware reasoning. 
 
Adding depth (RGB $\rightarrow$ RGB-D) yields no overall benefit and actively degrades collision risk assessment, with CR dropping from 52.6\% to 36.0\%. This confirms that depth data in mirror scenes can mislead models: the sensor reports distorted distances for reflected objects, and models that trust these values make worse judgments than those without depth at all. Adding the mirror mask (RGB $\rightarrow$ RGB-M) brings a modest gain on perception tasks, with MC improving from 39.8\% to 42.7\%, suggesting that explicitly marking reflective boundaries provides some useful context. However, combining both signals (RGB-DM) does not outperform the mask alone (RGB-M), as depth distortion offsets the benefit of mirror localization. Most notably, mirror-to-real spatial inference (MR) remains near floor at 2.1--3.2\% regardless of input mode. Even when depth data is available, models cannot leverage it for spatial reasoning in mirror scenes, reinforcing that the depth channel, rather than serving as a reliable geometric cue, becomes a source of misinformation when reflective surfaces are present.
Appendices~\ref{app:nonreflective_control} and~\ref{app:spatialrgpt_eval}
provide a control evaluation in non-reflective scenes and an
evaluation of a model with dedicated depth processing, respectively.
 
\begin{table}[t]
\centering
\caption{Impact of input modality on mirror-aware spatial reasoning. We report exact accuracy (\%) for each sub-task across four input configurations. RGB: raw image only; RGB-M: image with mirror region mask overlaid; RGB-D: image paired with depth map; RGB-DM: image with both depth map and mirror masks.}
\label{tab:input_modality}
\fontsize{8.5}{11}\selectfont
\fontsize{8.5}{10}\selectfont
\setlength{\tabcolsep}{2.5pt}
\renewcommand{\arraystretch}{1.0}

\begin{tabular*}{\linewidth}{@{\extracolsep{\fill}}l ccc cccc cccc c@{}}
\toprule
 & \multicolumn{3}{c}{Refl.-aware Perception} & \multicolumn{4}{c}{Spatial Reasoning} & \multicolumn{4}{c}{Scene-level Decision-making} & \\
\cmidrule(lr){2-4} \cmidrule(lr){5-8} \cmidrule(lr){9-12}
Input Mode & VC & MC & W-Avg & MR & CT & MPAC & W-Avg & RE & AS & CR & W-Avg & Overall \\
\midrule
\rowcolor{gray!15} \multicolumn{13}{l}{\textit{Qwen3.5-9B}} \\
\cmidrule{1-13}
RGB      & 32.2 & 39.8 & 34.6 & 2.1 & \textbf{20.9} & \textbf{52.7} & \textbf{20.6} & 57.2 & 8.1 & \textbf{52.6} & 29.8 & 30.3 \\
RGB-D    & 31.4 & 37.7 & 33.4 & 2.7 & 20.6 & 50.4 & 20.2 & 61.4 & 7.9 & 36.0 & 30.2 & 29.6 \\
RGB-M    & \textbf{32.6} & \textbf{42.7} & \textbf{35.8} & \textbf{3.2} & 19.7 & 49.8 & 20.0 & \textbf{62.1} & \textbf{10.3} & 45.7 & \textbf{32.4} & \textbf{31.3} \\
RGB-DM   & 32.5 & 42.4 & 35.6 & 2.9 & 17.2 & 49.8 & 19.0 & 56.0 & 9.8 & 38.3 & 29.3 & 30.4 \\
\bottomrule
\end{tabular*}
\end{table}

\subsection{Referring Protocol Analysis}
\label{sec:referring_analysis}

Prior work has shown that how target objects are referred to can significantly affect VLM performance~\cite{vrptest2023, vpbench2025}. To ensure that our findings reflect genuine mirror reasoning ability rather than artifacts of a particular referencing method, we compare five protocols on Qwen3.5-9B. Text-BBox and Text-BBox+Desc provide the original image with textual coordinate descriptions, the latter additionally including a natural language description of the target object. Visual-BBox draws bounding boxes directly on the image. Visual-Masks provides a binary mask of the target object alongside the original image. Visual-BBox+Desc combines drawn bounding boxes with textual descriptions. Examples of the five referring protocols are shown in
Figure~\ref{fig:referring_protocols} in Appendix~\ref{app:qa_construction}.

\begin{wraptable}{r}{0.42\linewidth}
\centering
\caption{Impact of object-referencing protocols on overall accuracy (\%). Results shown for Qwen3.5-9B with RGB-M input. Desc denotes textual object descriptions.}
\label{tab:referring}
\normalsize
\begin{tabular}{lc}
\toprule
Referring Protocol & Overall Acc. \\
\midrule
Text-BBox          & 28.0 \\
Text-BBox+Desc     & 31.6 \\
Visual-BBox        & \textbf{33.0} \\
Visual-Mask        & 27.9 \\
Visual-BBox+Desc   & \textbf{33.0} \\
\bottomrule
\end{tabular}
\end{wraptable}
 
As shown in Table~\ref{tab:referring}, overall accuracy ranges from
27.9\% to 33.0\%, a spread of 5.1 percentage points.
Visual bounding boxes achieve the best performance (33.0\%),
and adding textual descriptions on top provides no further gain.
For text-only referring, adding object descriptions improves
accuracy by 3.6 percentage points, indicating that natural language
context helps when visual cues are absent. Interestingly, binary
masks perform worst (27.9\%), possibly because models are less
accustomed to interpreting isolated mask overlays compared to
bounding boxes. The narrow overall range and the persistent gap
to human performance (85.0--90.7\%) confirm that object-referencing
method is not the primary bottleneck: regardless of how objects
are referred to, models still fail at mirror-aware reasoning itself.

\section{Discussion and Limitations}
While MIRA-Bench provides a first step toward evaluating reflection-aware spatial reasoning and covers common reflective scenarios, it remains infeasible to exhaustively cover all reflective materials, lighting conditions, mirror geometries, and multi-mirror configurations encountered in the real world. 
Furthermore, the current benchmark is based on static images, whereas real-world embodied systems often require reasoning over dynamic scenes, changing viewpoints, and temporal cues, for example in visual navigation tasks~\cite{10204881}.
Extending MIRA-Bench to dynamic scenes is an important direction for future work. Finally, although depth maps are included to examine RGB-D reasoning under reflection, incorporating depth information from other range sensors, such as LiDAR or sonar, could further improve the evaluation of VLMs’ scene understanding and geometric reasoning in reflective scenes. Appendix~\ref{app:depth_reliability} discusses how surface properties
and sensing technology affect depth reliability.
\section{Conclusion}
In this work, we introduced \textbf{MIRA-Bench}, the first evaluation benchmark designed to assess whether VLMs can perform spatial reasoning in scenes with reflective surfaces. MIRA-Bench addresses a critical blind spot in existing vision-language evaluation by providing grounded RGB-D annotations across diverse reflective scenarios, covering the reasoning pipeline from perception and spatial inference to scene-level decision-making. Through extensive benchmarking, we show that state-of-the-art VLMs exhibit persistent failures across multiple reasoning levels, with spatial inference through reflections proving especially challenging. Most notably, we find that depth information, widely treated as a reliable geometric cue, can become a source of misinformation in mirror scenes, degrading model performance rather than improving it. This finding challenges a foundational assumption in RGB-D spatial reasoning and highlights the need to rethink how multimodal models interpret sensor data in the presence of reflective surfaces. We hope MIRA-Bench will serve as a foundation for future research on reflection-aware spatial understanding in multimodal systems.   

\begin{ack}
This research is funded in part by ARC-Discovery grants
(DP220100800, DE230100477, and FT260100691 to XY).
Hao Yin acknowledges STSE support from The University of Queensland.
Yanbin Liu is supported by the Australian Government through an
ARC Discovery Early Career Researcher Award (DECRA),
project number DE260100509.
We also acknowledge support from the NVIDIA Academic Research Program.
We thank the anonymous reviewers and area chairs for their
constructive suggestions.
\end{ack}

\bibliographystyle{plain}
\bibliography{references}

\newpage
\appendix
\section*{Appendix}

\section{Dataset Construction and Quality Control}
\subsection{Annotators and Training}
\label{app:annotators_training}

The annotation team comprised five paid postgraduate annotators in computer science with prior experience in spatial question-answering annotation. Before annotating the dataset, each annotator completed eight hours of training, including annotation exercises covering all task types on a set of 100 images. This training established a shared understanding of the annotation procedures and question definitions. It also helped annotators resolve numerous detailed issues, including the delineation of mirror polygons to exclude mirror frames, the use of the camera coordinate frame for all spatial-relation judgments, the inclusion of mirrors in counting questions, and other related annotation details.

\begin{figure}[hp]
    \centering
    \includegraphics[width=1.0\linewidth]
        {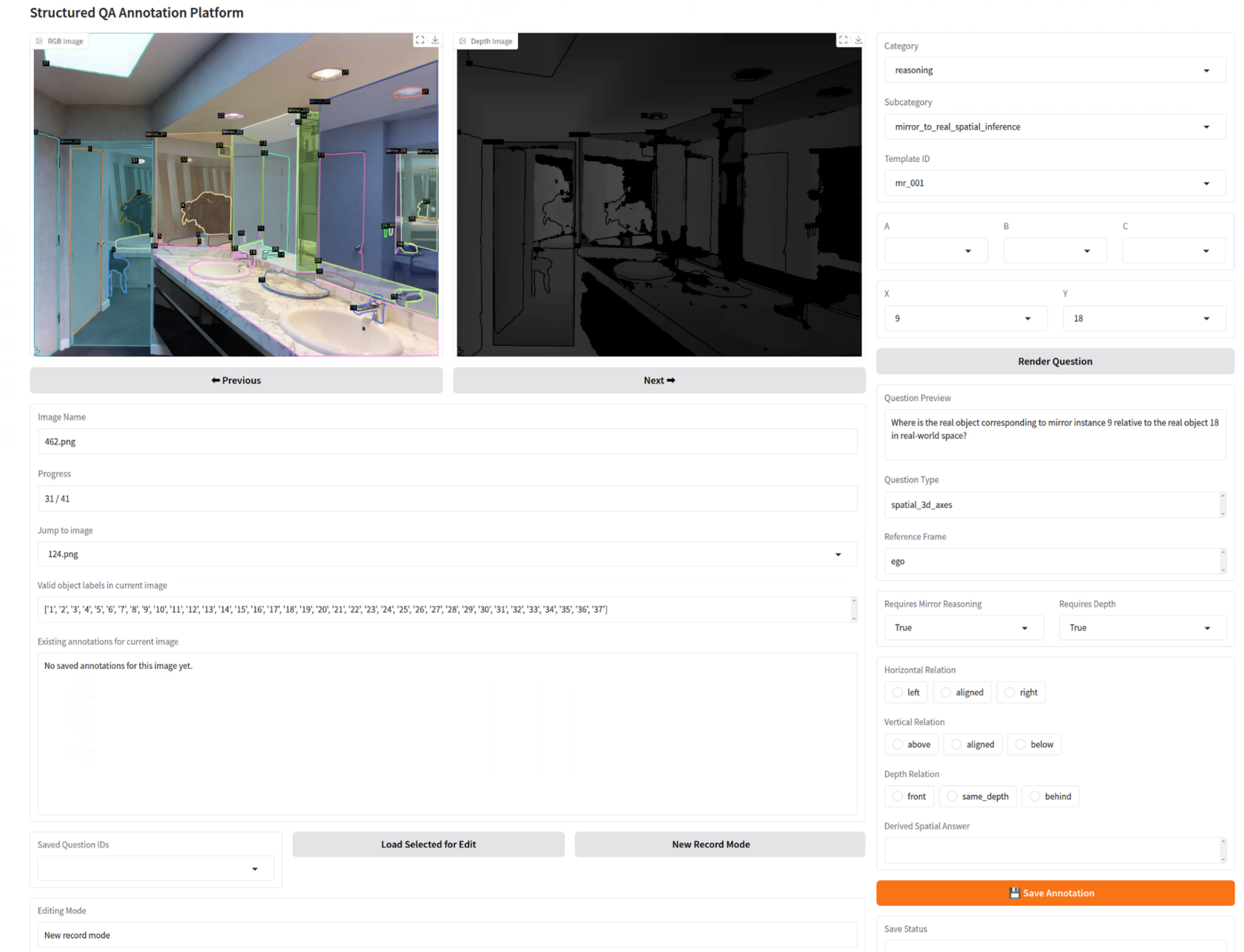}
    \caption{\textbf{Gradio-based QA annotation interface.} The interface displays the annotated RGB
    image and corresponding depth map, together with controls for task
    selection, instance referencing, question preview, and answer
    annotation. The example shows a mirror-to-real spatial-inference
    question with horizontal, vertical, and depth relation labels.}
    \label{fig:qa_annotation_interface}
\end{figure}

\paragraph{Annotation interfaces.}
In the first stage, we used X-AnyLabeling~\citep{wang2023xanylabeling}
to delineate reflective surface regions, annotate object instance
polygons, and record real--reflection correspondences. In the second
stage, we used a custom QA annotation interface built
with Gradio~\citep{abid2019gradio}. The interface (Figure~\ref{fig:qa_annotation_interface}) displayed the RGB image with region and instance annotations alongside the
corresponding depth map. Annotators could select task templates and
target instance identifiers, preview the instantiated questions, and
record answers. 

\subsection{Stage 1: Image Annotation and Quality Control}
\label{app:image_annotation}

For each image, the annotators delineated the reflective-surface regions
and annotated polygons for directly observed and reflected object
instances. Each instance was assigned a unique numeric identifier.
Visibility labels and real--reflection correspondences were also
recorded. These annotations provided the basis for object referencing
and subsequent QA construction.

\paragraph{Region and instance annotation audit.}
Reflective surface regions and object instance polygons were annotated
in batches of 100 images. Each batch was initially annotated by one
annotator. A second annotator then independently re-annotated a randomly
selected 20\% of the images without access to the initial annotations.
We computed the mean intersection over union (mIoU) between the two sets
of region and instance annotations on this subset.

Batches with an mIoU below 0.85 were reassigned to a third annotator, who
re-annotated the entire batch from scratch without access to either
previous set of annotations. Each reassigned batch underwent the same
audit procedure using a newly sampled 20\% subset and the same mIoU
threshold. The protocol permitted at most one reassignment per batch
and required joint review and adjudication of disagreements if the
second audit failed. In practice, all reassigned batches passed the
second audit.

\paragraph{Correspondence agreement.}
Inter-annotator agreement for real--reflection correspondences was
96.2\%. This assessment evaluated the consistency of correspondence annotations, complementing the mIoU-based evaluation of spatial overlap between region and instance annotations.

\subsection{Stage 2: QA Construction and Verification}
\label{app:qa_construction}

QA construction followed the image annotation stage. The generation
procedure differed across the three cognitive levels, as described
below.

\paragraph{Reflection-aware perception.}
Questions and answers were generated automatically from the annotated
visibility labels and real--reflection correspondences. For questions
requiring candidate instances, distractors were drawn from other annotated
instances in the same scene. Table~\ref{tab:mira-qa-perception} presents the
six templates for the visibility classification and mirror correspondence subtasks,
together with their response formats and candidate options.

\paragraph{Spatial reasoning and scene-level decision-making.}
Questions were generated from templates for each subtask using the
annotated instances. Response options were defined according to the subtask.
Annotators determined the answers by inspecting the RGB images and using
depth maps as a geometric reference. Tables~\ref{tab:mira-qa-reasoning} and~\ref{tab:mira-qa-decision}
present four question templates for spatial reasoning and four for
decision making at the scene level, respectively, together with their
response formats and answer options where applicable.
For the attribute completion subtask, when a textual description is
provided by the annotator, it is used to construct one answer option,
which is combined with distractors to form the final candidate set. In total, the annotation
tool defines 14 canonical question templates across eight subtasks.

\begin{table}[t]
\centering
\caption{\textbf{QA templates for reflection-aware perception.}
Braced placeholders denote instance identifiers.
\{X\} and \{Y\} denote objects visible in reflection or directly.
Candidate instances for the multiple-choice questions are drawn
from the same scene.}
\label{tab:mira-qa-perception}
\normalsize
\setlength{\tabcolsep}{5pt}
\renewcommand{\arraystretch}{1.10}
\begin{tabularx}{\linewidth}{@{}>{\raggedright\arraybackslash}p{1.70cm}>{\raggedright\arraybackslash}X@{}}
\toprule
\textbf{ID} & \textbf{Question template and response specification} \\
\midrule
\multicolumn{2}{@{}l}{\textbf{Visibility classification}} \\
\addlinespace[2pt]
\texttt{vc\_001} &
Is the object \{X\} visible in a mirror reflection but
not directly in this image?\newline
\textit{Format:} \texttt{yes\_no}.
\textit{Options:} yes; no. \\
\addlinespace[5pt]
\texttt{vc\_002} &
Is the object \{Y\} directly visible but not visible
in any mirror or other reflective surface in this image?\newline
\textit{Format:} \texttt{yes\_no}.
\textit{Options:} yes; no. \\
\addlinespace[5pt]
\texttt{vc\_003} &
Is the object \{X/Y\} visible both directly and
in a mirror reflection in this image?\newline
\textit{Format:} \texttt{yes\_no}.
\textit{Options:} yes; no. \\
\midrule
\multicolumn{2}{@{}l}{\textbf{Mirror correspondence}} \\
\addlinespace[2pt]
\texttt{mc\_001} &
Which real-world instance corresponds to mirror instance \{X\}?\newline
\textit{Format:} \texttt{single\_choice}.\newline
\textit{Options:} candidate real-world instance IDs. \\
\addlinespace[5pt]
\texttt{mc\_002} &
Which mirror instance corresponds to real-world instance \{Y\}?\newline
\textit{Format:} \texttt{single\_choice}.\newline
\textit{Options:} candidate mirror instance IDs. \\
\addlinespace[5pt]
\texttt{mc\_003} &
Do mirror instance \{X\} and real-world instance \{Y\}
correspond to the same real object?\newline
\textit{Format:} \texttt{yes\_no}.
\textit{Options:} yes; no. \\
\bottomrule
\end{tabularx}
\end{table}

\begin{table}[!h]
\centering
\caption{\textbf{QA templates for spatial reasoning.} Braced placeholders denote instance identifiers or spatial directions. \{X\} and \{Y\} denote objects visible in reflection or directly. For counting questions, annotators select \{DIR\}
from the six listed directions. Each template specifies the answer format and, where applicable, the response options and text field requirements.}
\label{tab:mira-qa-reasoning}
\normalsize
\setlength{\tabcolsep}{5pt}
\renewcommand{\arraystretch}{1.10}
\begin{tabularx}{\linewidth}{@{}>{\raggedright\arraybackslash}p{1.70cm}>{\raggedright\arraybackslash}X@{}}
\toprule
\textbf{ID} & \textbf{Question template and response specification} \\
\midrule
\multicolumn{2}{@{}l}{\textbf{Mirror-to-real spatial inference}} \\
\addlinespace[2pt]
\texttt{mr\_001} &
Where is the real object corresponding to mirror instance \{X\} relative to the real object \{Y\} in real-world space?\newline
\textit{Format:} \texttt{spatial\_3d\_axes}.\newline
\textit{Options:}\newline
\begin{tabular}[t]{@{\hspace{1em}}l@{}}
horizontal: left, aligned, right; \\
vertical: above, aligned, below; \\
depth: front, same\_depth, behind.
\end{tabular} \\
\addlinespace[5pt]
\texttt{mr\_003} &
Where is the real object corresponding to mirror instance \{X1\} relative to the real object corresponding to mirror instance \{X2\} in real-world space?\newline
\textit{Format:} \texttt{spatial\_3d\_axes}.\newline
\textit{Options:}\newline
\begin{tabular}[t]{@{\hspace{1em}}l@{}}
horizontal: left, aligned, right; \\
vertical: above, aligned, below; \\
depth: front, same\_depth, behind.
\end{tabular} \\
\midrule
\multicolumn{2}{@{}l}{\textbf{Counting-based spatial reasoning}} \\
\addlinespace[2pt]
\texttt{ct\_001} &
How many objects are there in the real world to the \{DIR\} of the real object corresponding to instance \{X/Y\}? (Count only polygon-annotated objects and the mirror itself. Do not count unannotated objects.)\newline
\textit{Format:} \texttt{count}; integer response.\newline
\textit{DIR options:} left; right; front; behind; above; below. \\
\midrule
\multicolumn{2}{@{}l}{\textbf{Mirror-provided attribute completion}} \\
\addlinespace[2pt]
\texttt{mpac\_001} &
Compared with the direct view of real object instance \{Y\}, does the mirror reveal any additional visible side or attribute of the same object? If yes, describe the additional information revealed through the mirror.\newline
\textit{Format:} \texttt{single\_choice\_plus\_text}.\newline
\textit{Options:}\newline
\begin{tabular}[t]{@{\hspace{1em}}l@{}}
A. Not sure.\\
B. No additional information is revealed.\\
C. The mirror shows only the same visible side as the direct view.\\
D. The mirror reveals an additional side but no distinct new appearance detail.\\
E. The mirror reveals an additional side with distinct new appearance information.\\
\textit{Text field:} free-text description if answer is E. \\
\end{tabular} \\
\bottomrule
\end{tabularx}
\end{table}

\begin{table}[!h]
\centering
\caption{\textbf{QA templates for scene-level decision-making.}
\{X\} and \{Y\} denote objects visible in reflection or directly. Braced placeholders denote instance identifiers or directions.
For collision-risk questions, annotators select \{DIR\}
from the six listed directions.
The response format and options are provided below each question template.}
\label{tab:mira-qa-decision}
\normalsize
\setlength{\tabcolsep}{5pt}
\renewcommand{\arraystretch}{1.10}
\begin{tabularx}{\linewidth}{@{}>{\raggedright\arraybackslash}p{1.70cm}>{\raggedright\arraybackslash}X@{}}
\toprule
\textbf{ID} & \textbf{Question template and response specification} \\
\midrule
\multicolumn{2}{@{}l}{\textbf{Reachability}} \\
\addlinespace[2pt]
\texttt{ac\_001} &
Is real-world instance \{X/Y\} reachable from the current camera position?\newline
\textit{Format:} \texttt{yes\_no}. \textit{Options:} yes; no; not sure. \\
\midrule
\multicolumn{2}{@{}l}{\textbf{Next-step action selection}} \\
\addlinespace[2pt]
\texttt{ac\_002} &
What is the best next action for the camera to move toward real-world instance \{X/Y\}?\newline
\textit{Format:} \texttt{single\_choice}.\newline
\textit{Options:} move left; move right; move forward; move backward; rotate left; rotate right; turn around; stop; not sure; none. \\
\addlinespace[5pt]
\texttt{ac\_003} &
How should the camera move to better observe the real object corresponding to mirror instance \{X\}?\newline
\textit{Format:} \texttt{single\_choice}.\newline
\textit{Options:} move left; move right; move forward; move backward; rotate left; rotate right; turn around; stop; not sure; none. \\
\midrule
\multicolumn{2}{@{}l}{\textbf{Collision-risk awareness}} \\
\addlinespace[2pt]
\texttt{ac\_004} &
Based on full-scene understanding, including real-world information inferred from both mirror and non-mirror regions, would moving the camera \{DIR\} now pose any collision risk?\newline
\textit{Format:} \texttt{yes\_no}. \textit{Options:} yes; no; not sure.\newline
\textit{DIR Options:} left; right; forward; backward; above; below. \\
\bottomrule
\end{tabularx}
\end{table}

\begin{figure}[tbp]
    \centering
    \includegraphics[width=\linewidth]{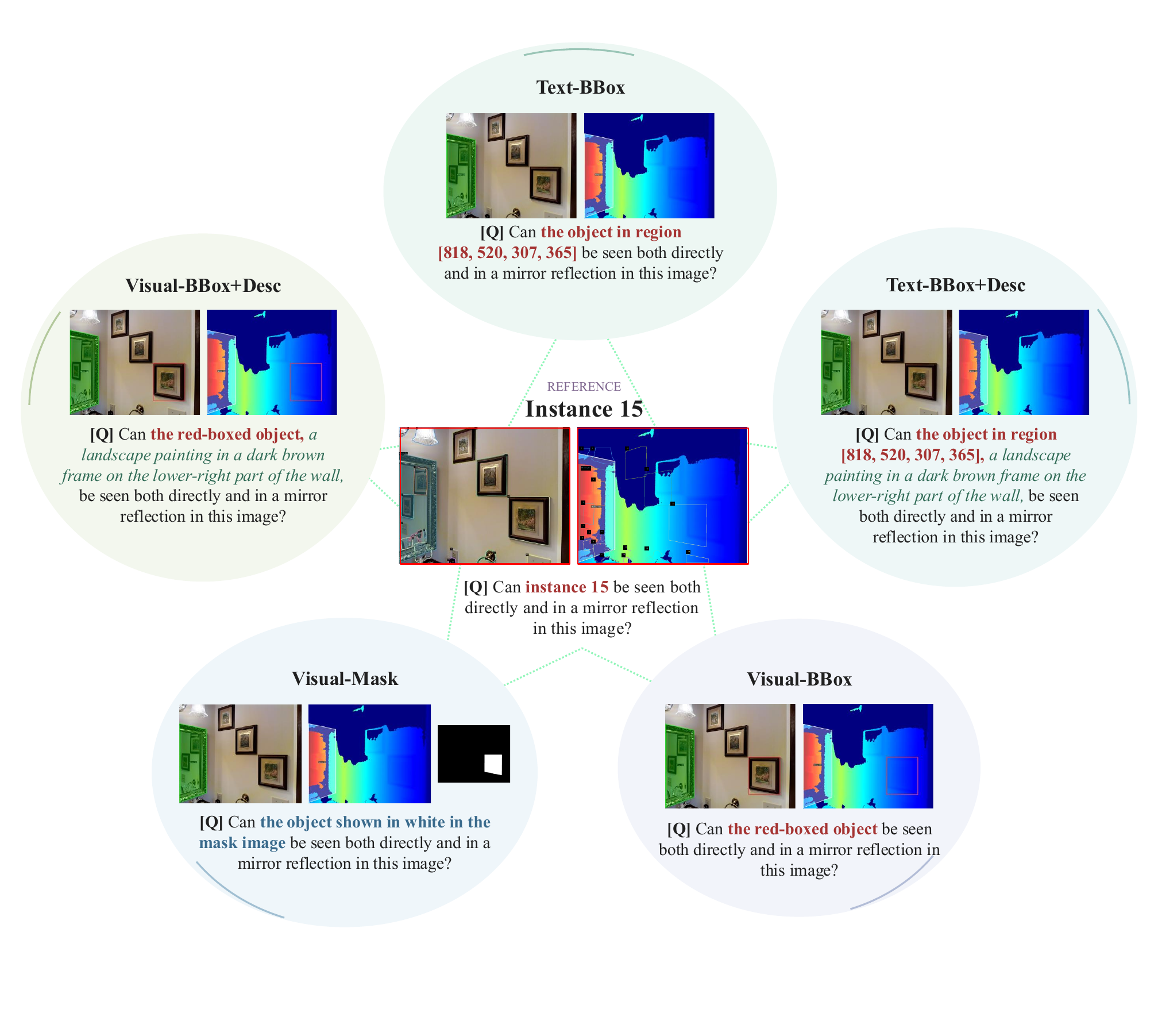}
    \caption{Examples of the five object-referencing protocols in
    MIRA-Bench. Each surrounding panel presents the same underlying
    visibility classification question for the same target object
    using a different referring protocol. The central panel identifies
    the target instance. RGB-DM inputs are shown for illustration.}
    \label{fig:referring_protocols}
\end{figure}

\paragraph{Object referencing.}
For each subtask, the resulting questions were formatted according to the applicable protocols for object referencing. These protocols identify the
relevant instances through text descriptions, bounding box coordinates,
visual bounding boxes, binary instance masks, or their supported
combinations. Questions about collision risk refer to a movement direction
rather than a target instance and therefore do not require an object
reference. Figure~\ref{fig:referring_protocols} illustrates the five protocols
using the same target instance and underlying question.

\paragraph{QA verification.}
Every QA pair, including those with automatically generated answers,
was cross-verified by two independent reviewers. The reviewers checked
question wording, answer correctness, and the plausibility of the
candidate options. Disagreements were resolved by a third reviewer.
Depth maps were consulted as additional geometric evidence for spatial
relations where applicable. Inter-annotator agreement for spatial
relationship labels was 93.4\%.

\begin{figure}[p]
  \centering
  \includegraphics[width=\linewidth]{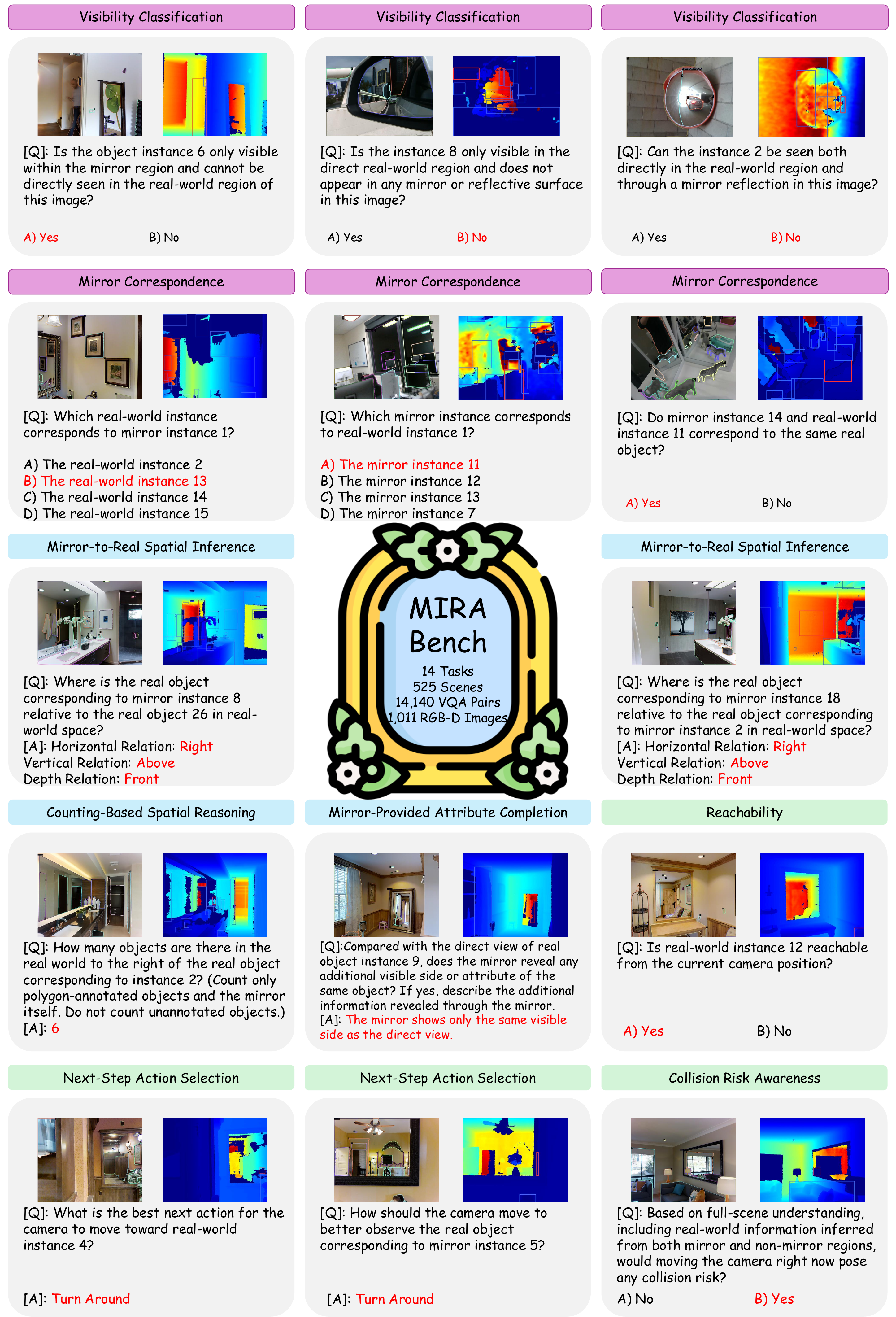}
  \caption{More visualization examples of MIRA-Bench.}
  \label{fig:more_visualizations}
\end{figure}

Figure~\ref{fig:more_visualizations} presents representative QA examples
from the three cognitive levels.

\clearpage
\section{Additional Analyses and Discussion}
\label{app:additional_analyses}

We present additional analyses of depth in non-reflective scenes,
cross-level error propagation, and the performance of
a model with dedicated depth processing. We then discuss how
surface properties affect depth reliability and contextualize
our findings.

\subsection{Depth in Non-Reflective Scenes}
\label{app:nonreflective_control}

To test whether performance decreases simply because an additional modality is
provided, we evaluate four models on 10,000 spatial relation QA samples from
the non-reflective 2D subset of RefSpatial~\citep{zhou2025roborefer}. The RGB and
RGB-D conditions use the same questions and evaluation protocol, differing only
in the availability of the depth map. Table~\ref{tab:refspatial_control} reports
the results.

\begin{table}[h!]
\centering
\caption{Accuracy (\%) on 10,000 non-reflective RefSpatial~\citep{zhou2025roborefer} questions. $\Delta$
denotes the accuracy difference (RGB-D minus RGB),
measured in percentage points.}
\label{tab:refspatial_control}
\small
\setlength{\tabcolsep}{10pt}
\renewcommand{\arraystretch}{1.05}
\begin{tabular}{@{}lrrr@{}}
\toprule
Model & RGB & RGB-D & $\Delta$ \\
\midrule
GPT-5.4          & 62.01 & 61.27 & $-0.74$ \\
Gemini-2.5-Flash & 62.50 & 61.98 & $-0.52$ \\
Qwen3.5-9B       & 66.17 & 65.62 & $-0.55$ \\
LLaVA-OV-7B      & 57.30 & 56.98 & $-0.32$ \\
\bottomrule
\end{tabular}
\end{table}

The absolute change is below one percentage point for every evaluated model in
this non-reflective control. By comparison, adding depth reduces Qwen3.5-9B's
collision-risk accuracy on MIRA-Bench from 52.6\% to 36.0\% (Table~\ref{tab:input_modality}). The control therefore argues against
an explanation based solely on the presence of an additional modality. It does
not imply that depth is uniformly harmful; rather, the larger decrease is
specific to the evaluated mirror-sensitive setting, in which the depth signal
can be physically misleading.

\subsection{Cross-Level Error Propagation Analysis}
\label{app:cross_level}

We analyze whether scene-level errors are associated with errors at lower
cognitive levels. We link
each reachability or action selection question to the visibility
classification and spatial inference questions concerning the same
target instance.
Collision-risk questions are excluded because they concern a movement direction
rather than a specific target instance.

\begin{table}[h!]
\centering
\caption{Scene-level failure rate (\%) grouped by the outcomes of linked
lower-level questions. The groups overlap except for ``both correct.''}
\label{tab:cross_level}
\small
\setlength{\tabcolsep}{10pt}
\renewcommand{\arraystretch}{1.05}
\begin{tabular}{@{}lrrr@{}}
\toprule
Lower-level outcome & GPT-5.4 & Gemini-2.5-Flash & Qwen3.5-9B \\
\midrule
Perception fails   & 68.3 & 65.8 & 73.4 \\
Reasoning fails    & 63.0 & 60.4 & 64.4 \\
At least one fails & 63.9 & 59.4 & 65.9 \\
Both correct       & 45.3 & 40.8 & 56.5 \\
\bottomrule
\end{tabular}
\end{table}

For all three models in Table~\ref{tab:cross_level}, the scene-level failure rate is higher when at least one linked
lower-level prediction is incorrect than when both are correct: 63.9\% versus
45.3\% for GPT-5.4, 59.4\% versus 40.8\% for Gemini-2.5-Flash, and 65.9\% versus
56.5\% for Qwen3.5-9B. Nevertheless, 40.8--56.5\% of linked scene-level
questions are still answered incorrectly when both lower-level predictions are correct.
Correct answers to the linked lower-level questions are therefore not sufficient
to ensure correct scene-level prediction. This analysis is correlational and covers
only scene-level questions that can be linked to  both perception and spatial reasoning questions.

\subsection{Depth-Specialized Model Evaluation}
\label{app:spatialrgpt_eval}

We additionally evaluate SpatialRGPT~\cite{spatialrgpt2024}, which incorporates depth information for grounded spatial reasoning. 
We evaluate this model and three general-purpose VLMs on the same
subset of questions referring to a single object instance as it supports a
single referred region per question.

\begin{table}[h!]
\centering
\caption{Accuracy (\%) on the common single-region question set. $\Delta$
denotes the accuracy difference (RGB-D minus RGB),
measured in percentage points.}
\label{tab:matched_depth_models}
\small
\setlength{\tabcolsep}{10pt}
\renewcommand{\arraystretch}{1.05}
\setlength{\fboxsep}{1pt}
\begin{tabular}{@{}lrrr@{}}
\toprule
Model & RGB & RGB-D & $\Delta$ \\
\midrule
GPT-5.4     & 54.47 & 55.04 & $+0.57$ \\
Qwen3.5-9B  & 49.96 & 48.38 & $-1.58$ \\
LLaVA-OV-7B & 29.77 & 29.16 & $-0.61$ \\
\colorbox[gray]{0.92}{SpatialRGPT} &
\colorbox[gray]{0.92}{37.48} &
\colorbox[gray]{0.92}{39.27} &
\colorbox[gray]{0.92}{$+1.79$} \\
\bottomrule
\end{tabular}
\end{table}

As shown in Table~\ref{tab:matched_depth_models}, adding depth increases its
accuracy from 37.48\% to 39.27\%. Among the other three models,
one shows a small improvement, while the remaining two show small
decreases in accuracy. These results indicate that depth provides
a modest benefit for this model, but does not consistently improve
performance across the evaluated models. Its accuracy remains low,
indicating persistent difficulties in reasoning about reflective scenes.

\subsection{Surface Properties and Depth Reliability}
\label{app:depth_reliability}
Depth sensing failures are not unique to mirrors and depend on both
surface properties and the sensing technology. Transparent objects
can produce missing measurements or depth values associated with
surfaces behind them rather than the objects themselves
~\cite{sajjan2019cleargrasp3dshapeestimation}. Translucent and specular surfaces can
likewise produce noisy, distorted, or incomplete depth
~\cite{wei2024d3roma}. In stereo estimation, complex light paths and
the mixing of foreground and background appearance can introduce
ambiguous correspondences, leading to incorrect estimates of
surface depth~\cite{wei2024d3roma}.

These failure modes contextualize the ambiguity of depth in the
reflective scenes studied by MIRA-Bench. A depth map may appear
locally plausible while failing to represent the relevant physical
surfaces or object locations. The benchmark therefore examines
whether VLMs can reason about reflective scenes when depth cannot
be assumed to provide reliable metric evidence. Our findings concern
the use of potentially misleading depth in spatial reasoning and
scene-level decision-making; they do not imply that depth is
intrinsically unreliable or uniformly detrimental to spatial reasoning.

\section{Computational Resources}
The experiments are conducted by using two NVIDIA GeForce RTX 4090 GPUs and Intel(R) Core(TM) i9-14900KF CPUs with 24 cores. The evaluation takes about 8 hours to complete for each model. 

\section{Societal Impacts}
This paper discusses an important but overlooked area in the MLLM benchmarks. It does not contain malicious use, privacy issues, and security issues. It does not contribute to the training of the DeepFake models.

\section{Licenses for Existing Assets}
RGBD-Mirror~\cite{pdnet2021} and DVMD-D~\cite{dvmd2023} are publicly available and remain subject to the applicable terms of use of their original sources. Our code was released under the Apache License 2.0, and our self-collected data were released under the CC BY-NC 4.0 license for non-commercial use.

\newpage
\section*{NeurIPS Paper Checklist}

\begin{enumerate}

\item {\bf Claims}
    \item[] Question: Do the main claims made in the abstract and introduction accurately reflect the paper's contributions and scope?
    \item[] Answer: \answerYes{} 
    \item[] Justification: see the Introduction Section.

\item {\bf Limitations}
    \item[] Question: Does the paper discuss the limitations of the work performed by the authors?
    \item[] Answer: \answerYes{} 
    \item[] Justification: The limitation section is included in this paper.

\item {\bf Theory assumptions and proofs}
    \item[] Question: For each theoretical result, does the paper provide the full set of assumptions and a complete (and correct) proof?
    \item[] Answer: \answerYes{} 
    \item[] Justification: see the benchmark construction section.

    \item {\bf Experimental result reproducibility}
    \item[] Question: Does the paper fully disclose all the information needed to reproduce the main experimental results of the paper to the extent that it affects the main claims and/or conclusions of the paper (regardless of whether the code and data are provided or not)?
    \item[] Answer: \answerYes{} 
    \item[] Justification: all code and data are fully provided. See the links.

\item {\bf Open access to data and code}
    \item[] Question: Does the paper provide open access to the data and code, with sufficient instructions to faithfully reproduce the main experimental results, as described in supplemental material?
    \item[] Answer: \answerYes{} 
    \item[] Justification: see the links.

\item {\bf Experimental setting/details}
    \item[] Question: Does the paper specify all the training and test details (e.g., data splits, hyperparameters, how they were chosen, type of optimizer) necessary to understand the results?
    \item[] Answer: \answerYes{} 
    \item[] Justification: The paper does not involve model training. All benchmark details are provided.

\item {\bf Experiment statistical significance}
    \item[] Question: Does the paper report error bars suitably and correctly defined or other appropriate information about the statistical significance of the experiments?
    \item[] Answer: \answerNo{} 
    \item[] Justification: We report model accuracies on the benchmark
without error bars, confidence intervals, or statistical significance tests.

\item {\bf Experiments compute resources}
    \item[] Question: For each experiment, does the paper provide sufficient information on the computer resources (type of compute workers, memory, time of execution) needed to reproduce the experiments?
    \item[] Answer: \answerYes{} 
    \item[] Justification: see the Experiments Compute Resources Section in Appendix.
    
\item {\bf Code of ethics}
    \item[] Question: Does the research conducted in the paper conform, in every respect, with the NeurIPS Code of Ethics \url{https://neurips.cc/public/EthicsGuidelines}?
    \item[] Answer: \answerYes{} 
    \item[] Justification: see the code links.

\item {\bf Broader impacts}
    \item[] Question: Does the paper discuss both potential positive societal impacts and negative societal impacts of the work performed?
    \item[] Answer: \answerYes{} 
    \item[] Justification: see the Societal Impacts Section in Appendix.
    
\item {\bf Safeguards}
    \item[] Question: Does the paper describe safeguards that have been put in place for responsible release of data or models that have a high risk for misuse (e.g., pre-trained language models, image generators, or scraped datasets)?
    \item[] Answer: \answerNA{} 
    \item[] Justification: the paper poses no such risks.

\item {\bf Licenses for existing assets}
    \item[] Question: Are the creators or original owners of assets (e.g., code, data, models), used in the paper, properly credited and are the license and terms of use explicitly mentioned and properly respected?
    \item[] Answer: \answerYes{} 
    \item[] Justification: see the Licenses for Existing Assets Section in Appendix.

\item {\bf New assets}
    \item[] Question: Are new assets introduced in the paper well documented and is the documentation provided alongside the assets?
    \item[] Answer: \answerYes{} 
    \item[] Justification: see the code and benchmark links.

\item {\bf Crowdsourcing and research with human subjects}
    \item[] Question: For crowdsourcing experiments and research with human subjects, does the paper include the full text of instructions given to participants and screenshots, if applicable, as well as details about compensation (if any)? 
    \item[] Answer: \answerNA{} 
    \item[] Justification: the paper does not involve crowdsourcing nor research with human subjects.

\item {\bf Institutional review board (IRB) approvals or equivalent for research with human subjects}
    \item[] Question: Does the paper describe potential risks incurred by study participants, whether such risks were disclosed to the subjects, and whether Institutional Review Board (IRB) approvals (or an equivalent approval/review based on the requirements of your country or institution) were obtained?
    \item[] Answer: \answerNA{} 
    \item[] Justification: the paper does not involve crowdsourcing nor research with human subjects.

\item {\bf Declaration of LLM usage}
    \item[] Question: Does the paper describe the usage of LLMs if it is an important, original, or non-standard component of the core methods in this research? Note that if the LLM is used only for writing, editing, or formatting purposes and does \emph{not} impact the core methodology, scientific rigor, or originality of the research, declaration is not required.
    \item[] Answer: \answerYes{} 
    \item[] Justification: Since our paper introduces a benchmark specifically for evaluating MLLMs/VLMs, the setup, prompting strategies, and inference usage of these models are central to our experiments. We explicitly describe how we queried these models and processed their outputs in Section 3 and Section 4.

\end{enumerate}

\end{document}